\documentclass{article}
\usepackage{spconf,amsmath,graphicx,booktabs,multirow}
\usepackage{url}
\usepackage{xcolor}

\title{TYPE-BALANCED FEDERATED LEARNING FOR VISUAL ANALOG METER READING}
\name{Weida Zhao, Logan Bellamy, Yazhou Tu, Jiaqi Wang\textsuperscript{1}\thanks{Corresponding author: jqwang@auburn.edu}}
\address{Department of Computer Science and Software Engineering, Auburn University, Auburn, USA}

\begin{document}
\ninept
\maketitle

\begin{abstract}
Analog dial meters are widely deployed in industrial application and utility sites, where environments and meter types vary and inspection data may be sensitive.
Currently, automatic meter readers must be individually developed and deployed for each environment and meter type in practice.
Deep learning could handle this variability but requires diverse labeled data that are costly to collect and update.  In practice, meter images are distributed across independent sites, each with limited labels, while raw images often cannot be pooled because of ownership, governance, or privacy constraints. To address these challenges, we present a federated framework for visual analog meter reading that enables multiple sites to collaboratively train a reading model without sharing their raw images.
Our framework consists of a four-stage pipeline: (1) dial localization (2) thin-structure segmentation
trained federatively across clients (3) polar unwrapping (4) tick-counting decoding for final reading. To enable systematic evaluation of this setting, we release \textbf{MeterFL}, a 1{,}382-image
mask-annotated dataset organized into deployment-motivated pseudo-clients
derived from visual attributes via deterministic rules, with dHash
near-duplicate control between the segmentation train and test splits. We
evaluate both segmentation quality and end-to-end reading accuracy. MeterFL is publicly available at \url{https://github.com/weidazhaoooo/Meter-FL}.

\end{abstract}


\section{Introduction}
\label{sec:intro}

Dial-based meters remain widely used in utilities, factories, and medical facilities~\cite{salomon2020deep,reitsma2024pressure}. Automatic reading can significantly reduce the manual inspection effort required for system maintenance. However, training computer-vision-based reading methods that rely on semantic segmentation requires pixel-level annotations of thin pointers and individual scale ticks~\cite{wang2024corruption}. Producing such fine-grained masks is labor-intensive, while each deployment site may contain only a limited variety of meter types and imaging conditions. Consequently, an individual site may have only a small labeled dataset with limited visual diversity~\cite{leonalcazar2024gauges,wang2025keypoint}. In practice, data from different sites can be complementary, with each site contributing different meter types and imaging conditions. Sharing such data could help build a larger and more comprehensive training dataset~\cite{kairouz2021advances}. However, data-ownership and privacy requirements may prevent sites from sharing raw inspection images, making it difficult to construct a dataset that covers a broad range of meter types and imaging conditions~\cite{nikic2024digit,kairouz2021advances}.
Federated learning
(FL)~\cite{mcmahan2017fedavg} enables collaborative training by exchanging
model updates while keeping images local~\cite{kairouz2021advances}.
Applying FL alone does not resolve heterogeneity in instrument types,
imaging conditions, and labeled-data volumes~\cite{li2020fedprox,li2021fedbn}. Standard FedAvg assigns
aggregation weights proportional to sample counts, so data-rich types
receive most of the weight under severe type imbalance~\cite{mcmahan2017fedavg}. This may limit
performance on underrepresented instruments~\cite{li2020qffl}. Uniform client weights do
not necessarily balance types either, since some types span more clients
than others. We target this type imbalance so that collaboration better
serves both common and less frequent instruments.
Studying this problem also requires a federated benchmark. Existing
meter-reading benchmarks focus on centralized learning or reading
accuracy~\cite{salomon2020deep,measurebench,dialbench}; to our knowledge,
no established benchmark provides a common protocol for federated analog
pointer-meter reading. Reproducible client partitions and evaluation of
both segmentation and downstream readings are needed to assess
aggregation under client heterogeneity~\cite{lai2022fedscale}.

Our contributions are  \textbf{(1)}~\textbf{MeterFL}, a 1{,}382-image
mask-annotated benchmark with attribute-derived pseudo-clients and
near-duplicate-controlled train/test splits; \textbf{(2)}~a four-stage meter-reading framework
with \emph{type-balanced aggregation} for tick and pointer segmentation,
assigning equal total weight to each instrument type and sample-size
weights within types; and \textbf{(3)}~evaluation
across three initializations and five federated baselines. Type balancing aggregation, which assigns equal total aggregation weight to each instrument type while weighting clients within each type by sample size
improves average pointer intersection-over-union (IoU) over FedAvg and isolated local training
across all three initializations, and reduces cross-client IoU variation
relative to FedAvg. With synthetic initialization, it achieves the
highest average pointer IoU (0.576) and lowest client STD (0.153) among
the compared federated methods.

\begin{figure*}[t]
\centering
\includegraphics[width=\textwidth]{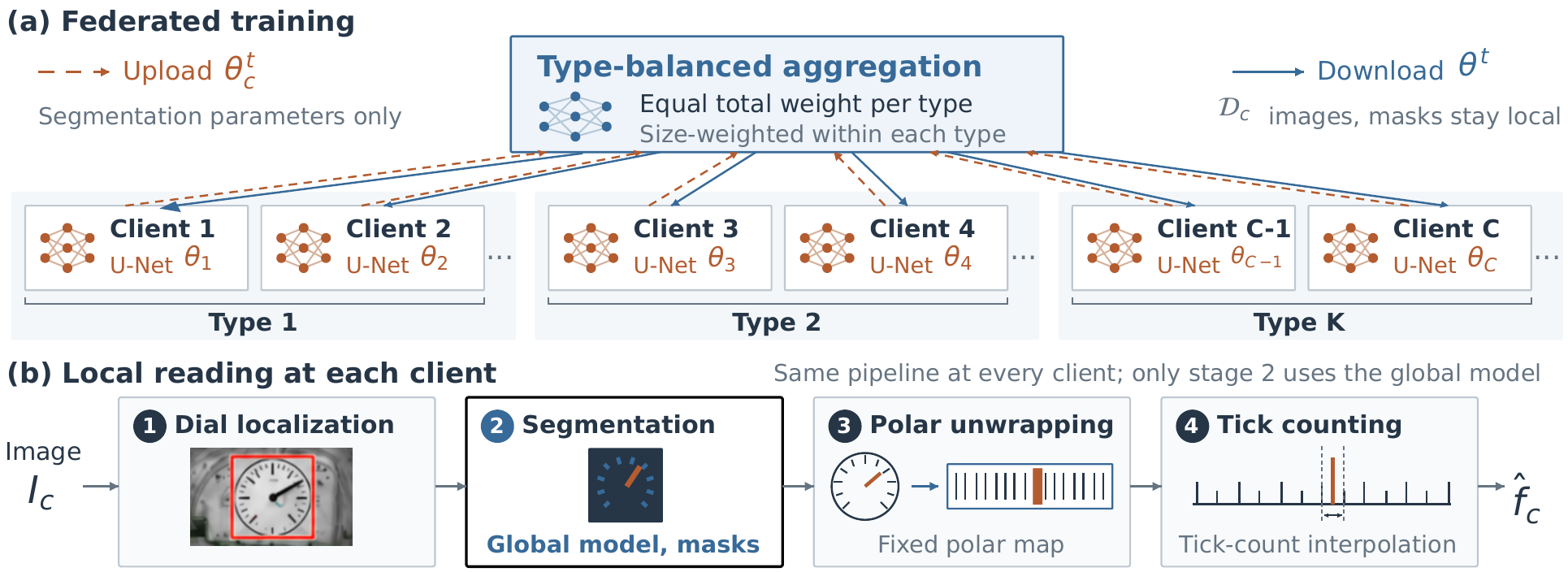}
\caption{Federated meter-reading framework. (a)~Clients exchange only
segmentation parameters with a type-balanced server: each instrument type
receives equal total weight, with sample-size weighting within types.
(b)~The shared local reading pipeline, with schematic stage illustrations
and solid inference arrows. The gray dotted guide identifies stage 2,
the only federatively trained component. Images and masks remain local. 
}
\label{fig:pipe}
\end{figure*}

\section{Related Work}
\label{sec:related}

\textbf{Analog meter reading.} Conventional methods combine dial
localization, pointer detection, scale extraction, and geometric
interpolation~\cite{lauridsen2019reading}. Learning-based readers replace
handcrafted visual stages with detection, segmentation, or keypoint
networks~\cite{salomon2020deep,zhao2023substation,zhao2023identification,wang2025keypoint}.
Leon-Alcazar et al.~\cite{leonalcazar2024gauges} study synthetic-data
training for analog gauges; we treat this as an optional initialization,
not the aggregation mechanism. Vision-language models (VLMs) offer an
alternative reading route: MeasureBench~\cite{measurebench} evaluates
visual measurement reading, while DialBench~\cite{dialbench} targets
pointer meters and incorporates pointer--scale relations in fine-tuning.
Our pipeline retains explicit tick and pointer geometry and is evaluated
separately against these VLM-based readers.

\noindent\textbf{Federated visual metering.} FL enables collaborative visual
learning across institutions without pooling raw images, as demonstrated
in medical imaging~\cite{sheller2020federated,rieke2020future}.
For camera-based metering, Niki\'c et al.~\cite{nikic2024digit} apply FL
to digit recognition on numeric digital meter registers. Our task instead
requires dense tick/pointer segmentation and geometric decoding of a
continuous analog reading, with heterogeneous instrument types across
clients.

\noindent\textbf{Aggregation under client heterogeneity.} FedAvg aggregates
local models using sample-size weights~\cite{mcmahan2017fedavg}.
FedProx constrains local updates with a proximal
term~\cite{li2020fedprox}; FedBN retains client-specific normalization
to address feature shift~\cite{li2021fedbn}; and FedSAM uses
sharpness-aware local optimization~\cite{qu2022fedsam}.
q-FFL emphasizes high-loss clients through a fairness-oriented
objective~\cite{li2020qffl}. Our rule instead balances declared
instrument types: each type receives a fixed share of the aggregate,
independent of its data volume or number of clients.

\section{Methodology}
\label{sec:method}

\subsection{Meter-Reading Pipeline}
\label{sec:pipeline}

\begin{figure}[t]
\centering
\includegraphics[width=\columnwidth]{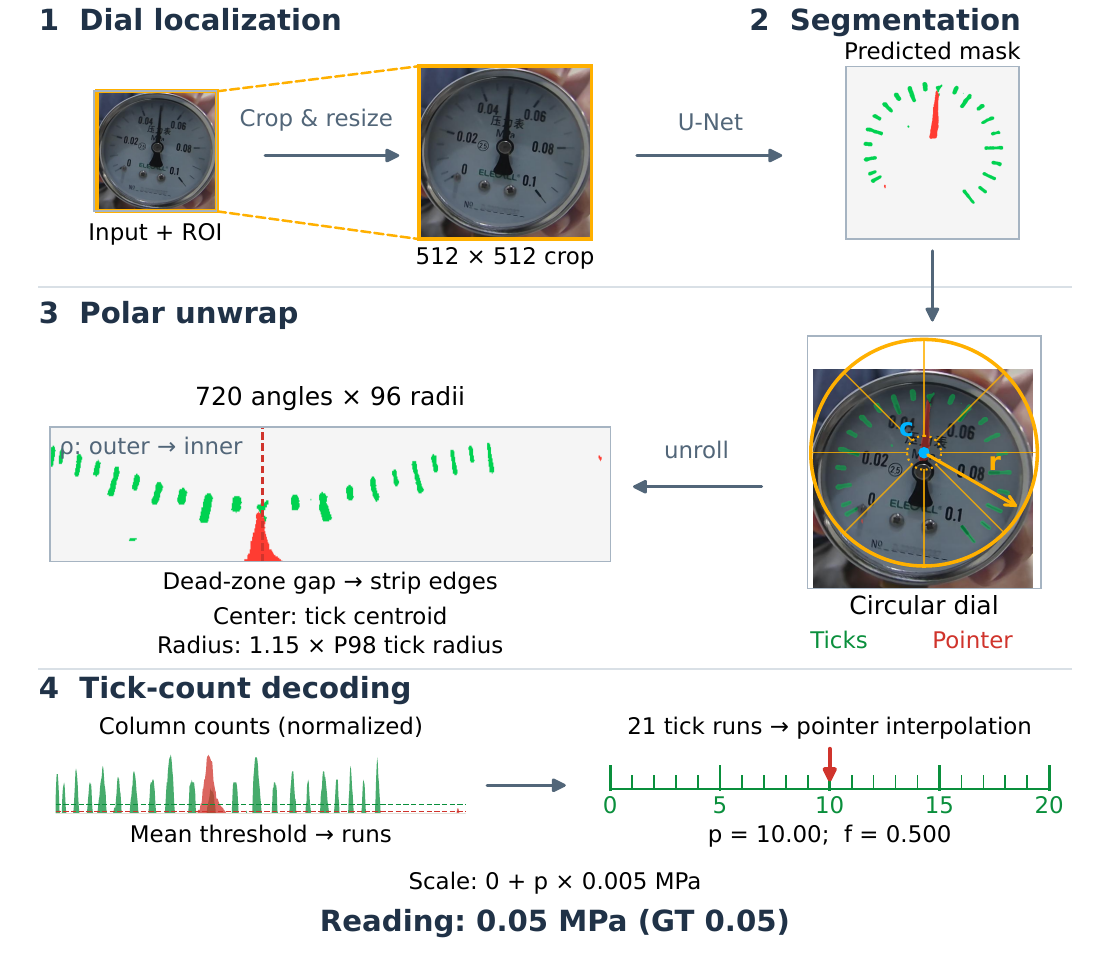}
\caption{Four-stage reading of a real meter image:
(1)~the orange ROI is cropped and resized, shown enlarged relative
to the close-up input;
(2)~tick and pointer segmentation;
(3)~circular sampling and polar unwrapping;
(4)~tick-count decoding.
The pointer is 10 of the 20 tick intervals from zero
($\hat f=0.5$). Each interval represents 0.005\,MPa, giving a
reading of $10 \times 0.005 = 0.05$\,MPa, which matches the ground truth.}
\label{fig:unroll}
\end{figure}

Each client independently runs the same four-stage meter-reading
pipeline (Fig.~\ref{fig:pipe}): (1)~dial localization; (2)~semantic
segmentation of ticks and pointer; (3)~polar unwrapping into a linear
tick ruler; and (4)~tick-count decoding into a reading. Each client keeps its images and annotations locally and shares only model updates with the server. 
The stage-1 detector is trained separately and then kept fixed throughout all subsequent experiments.
Stage 3 applies a fixed polar-unwrapping procedure, and stage 4 converts the unwrapped masks into readings using deterministic tick counting and interpolation. Neither stage contains trainable parameters. Collaborative training of stage 2 is described separately in Sec.~\ref{sec:aggregation}.
Fig.~\ref{fig:unroll} illustrates the four stages on a real meter image example.

\noindent\textbf{Stage 1: dial localization.}
A YOLO detector localizes the dial region and outputs a bounding box $b$.
This step reduces the area processed by the stage 2 segmentation network by removing most of the irrelevant background.
The detected region is slightly expanded by a factor $\gamma > 1$ to preserve the full dial face, converted to a square crop, and resized to $S \times S$ as the input to stage 2.
Here, $\gamma$ controls the amount of surrounding context retained, and $S$ sets the input resolution of the segmentation network.

\noindent\textbf{Stage 2: thin-structure segmentation.} We use a standard
U-Net~\cite{ronneberger2015unet} backbone with a pretrained CNN encoder
and skip decoder to map the dial crop to a three-class mask of
\emph{ticks}, \emph{pointer}, and \emph{background}. The predicted tick
and pointer masks provide the geometry for the following two stages.

\noindent\textbf{Stage 3: polar unwrapping.}
We use the centroid of the predicted tick pixels as the dial center.
The outer radius is set to $\beta r_q$, where $r_q$ is a high percentile of the distances from the tick pixels to the estimated center, and $\beta > 1$ adds a small margin to avoid cropping the outer ticks.
The predicted mask is then transformed into a polar strip with $A$ angular columns and $R$ radial rows, converting the circular scale into a linear representation.
We choose the largest gap between neighboring ticks as the start-end boundary of the unwrapped strip.

\noindent\textbf{Stage 4: tick-counting decoder.} For each angular column of the
unwrapped strip, we count the predicted tick and pointer pixels to obtain
two 1-D signals. Each signal is thresholded by its mean value, and
consecutive columns above the threshold are grouped into one detected
tick or pointer. The center of each group gives its angular position.
Let $\theta_i$ denote the center of tick $i$ and $\theta_{\rm ptr}$ the
pointer center. If the pointer lies between two adjacent ticks $i$ and
$i+1$, its position in tick intervals is
\begin{equation}
p = i + \frac{\theta_{\rm ptr}-\theta_i}{\theta_{i+1}-\theta_i},
\qquad
\hat f = \frac{p}{n-1},
\end{equation}
where $n$ is the number of detected ticks, $p\in[0,n-1]$ is the pointer
position measured in tick intervals, and $\hat f\in[0,1]$ is the
normalized gauge reading.
The scale interval $\Delta v$
and minimum value $v_{\min}$ of the gauge model are needed:
\begin{equation}
\hat v = v_{\min} + p\,\Delta v.
\end{equation}

\subsection{Type-Balanced Federated Aggregation}
\label{sec:aggregation}

Our aggregation method prevents data-rich instrument types from
dominating the global update by giving each type equal total weight.
The FL server connects the parallel clients' segmentation models,
exchanging only their parameters while the rest of each pipeline
remains local.

\noindent\textbf{Local training.} At round $t$, every client $c$ receives the global parameters
$\theta^{t-1}$, runs $E$ local epochs on its private images to obtain
$\theta_c^{t}$, and uploads parameters only. To handle sparse foreground
pixels, local training uses cross-entropy with inverse-frequency class
weights $w_k\propto 1/f_k$ clipped to $[w_{\min},w_{\max}]$, photometric
augmentation, and light Gaussian input noise.

\noindent\textbf{Type-balanced aggregation.}
We assign equal total aggregation weight to each instrument type.
Within each type, clients receive weights proportional to their
training-set sizes. The server aggregates the client models as
\begin{equation}
\theta^{r}=\sum_c w_c\theta_c^{r},
\qquad
w_c=\frac{1}{|\mathcal T|}
\frac{n_c}{\sum_{c':\,t(c')=t(c)}n_{c'}},
\end{equation}
where $\mathcal T$ is the set of instrument types, $t(c)$ is the
type of client $c$, and $n_c$ is its number of training images.
Thus, each type contributes $1/|\mathcal T|$ of the aggregated
model, regardless of how many clients or images it contains.
The server broadcasts the aggregated model for the next round.

\noindent\textbf{Optional synthetic initialization.} Federated training can start
from a model pretrained on procedurally rendered gauges whose masks are
generated automatically, so the prior costs no manual labels. It raises
average IoU for three of the four training schemes (Sec.~\ref{sec:exp}).
We evaluate this initialization as a separate factor.

\section{Experiments}
\label{sec:exp}

\subsection{The MeterFL Benchmark}
\label{sec:data}

\begin{figure}[t]
\centering
\includegraphics[width=\columnwidth]{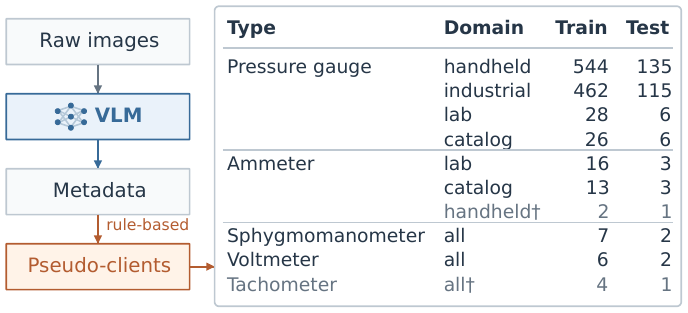}
\caption{Metadata-driven construction of MeterFL (left) and the resulting
pseudo-clients with train/test image counts (right). VLM descriptions are
mapped to pseudo-clients by deterministic rules. $^{\dagger}$Below the
minimum-size gate (gray).}
\label{fig:benchmark}
\end{figure}

We propose MeterFL, a benchmark for federated meter reading. Its
1{,}382 images come from the internet and two open-source meter-reading
datasets~\cite{measurebench}; every image carries polygon masks for
\emph{ticks}, \emph{pointer}, and \emph{background}, and a reading
ground truth for accuracy evaluation. Since the source dataset has no natural client identities, we propose a VLM-based pseudo-client construction method (Fig.~\ref{fig:benchmark}).
Based on our application scenario, a VLM annotator describes each image using seven visual-domain attributes, including instrument type, viewpoint, crop level, background, image quality, artifacts, and acquisition style, under a closed vocabulary and based only on visible evidence. An ordered first-match rule list then maps two of these attributes, acquisition style and background, to one of four acquisition domains. Product or web photos and images with clean white backgrounds are assigned to \emph{catalog}, industrial photos or images showing industrial scenes to \emph{industrial}, laboratory photos to \emph{lab}, and all remaining images to \emph{handheld}. A client is defined as an instrument$\times$domain cell. Cells with fewer than eight images are merged into the corresponding instrument's handheld cell, while an instrument represented by only a single cell forms one client. This procedure yields 10 pseudo-clients, including four large pressure-gauge clients (Fig.~\ref{fig:benchmark}, right).\begin{table*}[t]
\centering
\setlength{\tabcolsep}{1.6mm}
\renewcommand{\arraystretch}{0.92}
\resizebox{0.8\linewidth}{!}{
\begin{tabular}{@{}llccccc@{}}
\toprule
& & \multicolumn{2}{c}{Segmentation} & \multicolumn{3}{c@{}}{Meter reading} \\
\cmidrule(lr){3-4}\cmidrule(l){5-7}
Training & Method & Avg IoU & IoU STD & Med.\ \%FS & $\leq$5\%FS & Acc \\
\midrule
\multirow{9}{*}{Centralized}
 & Frontier VLM (zero-shot) & -- & -- & 18.0 & 7.4 & 49.1 \\
 & Qwen2.5-VL-7B (zero-shot) & -- & -- & 18.4 & 9.7 & 35.3 \\
 & Qwen2.5-VL-32B (zero-shot) & -- & -- & 17.2 & 20.1 & 21.9 \\
 & Qwen2.5-VL-72B (zero-shot) & -- & -- & 13.4 & 13.4 & 45.4 \\
 & InternVL3.5-8B (zero-shot) & -- & -- & 13.5 & 17.1 & 51.7 \\
\cmidrule(l){2-7}
 & LoRA-SFT, Qwen2.5-VL-7B (pooled 122 images) & -- & -- & 2.1 & 69.9 & 76.2 \\
 & PRI-style SFT + dial crop (pooled 122 images) & -- & -- & 2.4 & 70.6 & 77.0 \\
 & U-Net, pooled images (upper bound for FL) & 0.542$\pm$.02 & 0.184$\pm$.01 & \textbf{1.24} & 82.8 & 87.9 \\
 & U-Net, pooled images, type-balanced sampling & \textbf{0.569$\pm$.01} & \textbf{0.144$\pm$.00} & 1.27 & \textbf{84.8} & \textbf{88.2} \\
\midrule
\multirow{6}{*}{Federated}
 & FedAvg (size weights)~\cite{mcmahan2017fedavg} & 0.451$\pm$.01 & 0.238$\pm$.00 & 1.31 & 80.4 & 85.4 \\
 & FedBN~\cite{li2021fedbn} & 0.430$\pm$.01 & 0.234$\pm$.01 & 1.32 & 82.2 & 85.0 \\
 & FedProx ($\mu{=}.01$)~\cite{li2020fedprox} & 0.355$\pm$.03 & 0.162$\pm$.01 & 2.18 & 76.1 & 79.8 \\
 & FedSAM ($\rho{=}.05$)~\cite{qu2022fedsam} & 0.439$\pm$.01 & 0.236$\pm$.00 & 1.40 & 81.5 & 85.6 \\
 & q-FFL-style ($q{=}1$)~\cite{li2020qffl} & 0.475$\pm$.00 & 0.226$\pm$.00 & \textbf{1.29} & 80.4 & 85.7 \\
 & \textbf{Type-balanced (ours)} & \textbf{0.576$\pm$.01} & \textbf{0.153$\pm$.01} & 1.43 & \textbf{84.1} & \textbf{87.2} \\
\bottomrule
\end{tabular}}
\caption{Centralized versus federated training on MeterFL: average and
population STD of per-client pointer IoU (eight clients, 272 test images;
mean$\pm$std over 3 seeds) and median \%FS, share within 5\%FS, and Acc
(269 images; means over 3 seeds). VLMs have no segmentation columns.
Bold: best per regime.}
\label{tab:fl}
\end{table*}
To prevent train-test leakage from near-duplicate images, we use difference hashing (dHash), a perceptual hash based on neighboring-pixel intensity differences, to identify visually similar images by their hash distance. Images from the same near-duplicate group are never split across the training and test sets.

\subsection{Experimental Setup}
\label{sec:setup}

We use the MeterFL benchmark (Sec.~\ref{sec:data}) for training and
evaluation. Clients with fewer than five training or two test images
(tachometer, handheld ammeter; Fig.~\ref{fig:benchmark}) are excluded.
The remaining eight clients each train on all of their labeled images
(6--544 per client, 1{,}102 in total) for 60 rounds of 5 local epochs
(Adam $10^{-3}$, batch 16) over three seeds, so local computation grows
with client size as in FedAvg. All pipeline parameters are fixed in
advance: dial crops use $\gamma{=}1.4$ and $S{=}512$; the segmentation
network is a U-Net with an ImageNet-pretrained ResNet50 encoder and a
batch-normalized skip decoder (30.6M parameters), trained with class
weights clipped to $[0.02,30]$; the unwrap uses $\beta{=}1.15$ at the
98th-percentile tick radius and $A\times R=720\times96$. Unless stated
otherwise, models start from the synthetic initialization.
Segmentation is scored on the 272 test images of the eight clients by
average pointer IoU (the mean of per-client IoUs) and client IoU STD
(their population standard deviation). Complete-pipeline reading is
scored on the 269 GT-decodable images by median \%FS, the error
$|\hat f_{\rm pred}-\hat f_{\rm gt}|\times100$ against the
original-resolution GT-mask decode, the share of readings within
5\%FS, and Acc, which accepts errors within two minor divisions of
human readings under model-blind anchor calibration (median division
3.1\% of span).

\begin{figure}[t]
\centering
\includegraphics[width=\columnwidth]{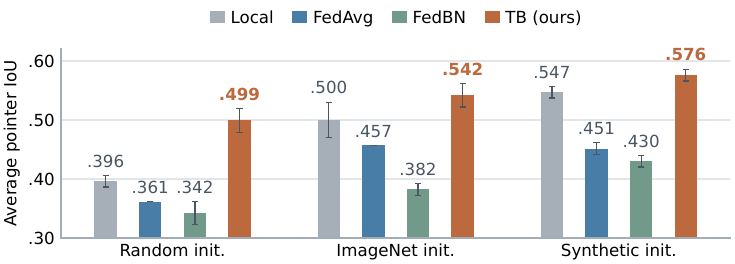}
\caption{Effect of initialization on average pointer IoU over the 272
test images of the eight clients. Bars show means over three
seeds; error bars show seed standard deviations.}
\label{fig:init}
\end{figure}

\subsection{Main Results: Centralized versus Federated Training}
\label{sec:fl}

Table~\ref{tab:fl} compares centralized learning, federated learning,
and our framework. In centralized learning we compare with centralized
VLM readers, both zero-shot and fine-tuned on pooled images; in
federated learning we compare with different aggregation methods.
Zero-shot VLMs reach only 13.4--18.4\%FS median; after fine-tuning on
a pooled 122-image subset, LoRA-SFT reaches 2.1\%FS and a PRI-style
variant after DialBench~\cite{dialbench} 2.4\%FS and 77.0\% Acc. Our
framework, whose only learned component is the 30.6M-parameter U-Net,
trained federatively without pooling any image, reaches 1.4\%FS and
87.2\% Acc. A small segmentation model with explicit tick geometry
thus beats 7B-parameter fine-tuned VLMs that see all the images.

Among federated methods, FedProx underfits (0.355 IoU); FedSAM and
FedBN spread across clients as much as FedAvg does (STD 0.234--0.236
vs.\ 0.238); q-FFL narrows the spread only slightly (0.226). Balancing
by instrument type reaches 0.576 IoU and 0.153 STD, the best in both
columns, at no extra communication. The reason is the weights
themselves: the four pressure-gauge clients hold 96\% of the FedAvg
aggregate but 25\% under type balance.

Centralized training on the pooled images is the upper bound for any
federated method: 87.9\% Acc, or 88.2\% with type-balanced sampling.
TB reaches 87.2\% and FedAvg 85.4\%. This difference is not
statistically significant (paired 95\% CI includes zero), but TB's
share of readings within 5\%FS is significantly higher (84.1 vs.\
80.4\%; CI $[+0.3, +7.3]$). Reading accuracy
therefore changes much less than segmentation does; our claim is better
segmentation and cross-client balance, with reading accuracy kept.

\subsection{Ablations and Sensitivity}
\label{sec:abl}

\begin{figure}[t]
\centering
\includegraphics[width=\columnwidth]{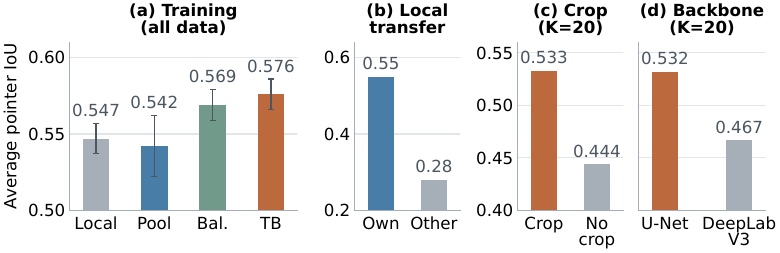}
\caption{Pointer-IoU controls. (a)~Full-data training; Pool/Bal.:
centralized training without/with type-balanced sampling.
(b)~Local-model transfer. (c,d)~$K{=}20$ TB controls with synthetic
and ImageNet initialization, respectively. Error bars: reported seed STD.}
\label{fig:controls}
\end{figure}

We evaluate initialization (Fig.~\ref{fig:init}) and the remaining
controls (Fig.~\ref{fig:controls}) using average pointer IoU.

\noindent\textbf{Initialization} (Fig.~\ref{fig:init}). TB achieves the highest
pointer IoU under all three initializations, 0.029--0.103 above the
next method. End-to-end Acc varies less (83.6--87.2\% across all
methods and initializations), so the gain lies mainly in segmentation.

\noindent\textbf{Aggregation controls} (Fig.~\ref{fig:controls}a,b).
Type-balanced sampling improves pooled training, and TB reaches a similar
average IoU without pooling images. The transfer control separately
shows that local models perform worse on other clients than on their own.
Under the $K{=}20$ protocol, retaining the dial crop and using U-Net
improve pointer IoU over their respective alternatives. 

\section{Conclusion}
\label{sec:concl}

We presented We developed a type-balanced aggregation based federated framework for visual analog meter reading
Experiments show that type-balanced aggregation consistently improves pointer segmentation over FedAvg and local training while reducing cross-client performance disparity, and achieves performance close to centralized training without sharing client images. This work explores the potential of FL for meter reading and provides a potential method to scale across distributed deployments.

\clearpage

\bibliographystyle{IEEEbib}
\bibliography{refs}

\end{document}